\def\year{2021}\relax
\documentclass[letterpaper]{article} 
\usepackage{aaai21}  
\usepackage{times}  
\usepackage{helvet} 
\usepackage{courier}  
\usepackage[hyphens]{url}  
\usepackage{graphicx} 
\usepackage{natbib}  
\usepackage{caption} 
\usepackage{amsmath}
\DeclareMathOperator*{\argmax}{argmax}
\usepackage{booktabs}
\usepackage{xcolor}
\usepackage{soul}
\usepackage[utf8]{inputenc}
\usepackage{amsfonts}
\usepackage{algorithm}
\usepackage{algorithmic}
\usepackage{amssymb}
\usepackage{flushend}
\usepackage{stfloats}
\usepackage{subfigure}
\usepackage{url}
\usepackage{epstopdf}
\usepackage{multicol}
\usepackage{dsfont,amsfonts,amssymb,color}
\usepackage{booktabs}
\usepackage{multirow}
\usepackage{fixltx2e}

\renewcommand{\footnotesize}{\normalsize}

\def\0{{\bf 0}}
\def\1{{\bf 1}}

\def\LM{{\mathcal L}}

\def\argmax{\mathop{\rm argmax}}

\title{Fair Representation Learning for Heterogeneous Information Networks}
\author{Ziqian Zeng$^2$, Rashidul Islam$^1$, Kamrun Naher Keya$^1$, James Foulds$^1$, Yangqiu Song$^2$, Shimei Pan$^1$\\
}
\affiliations{
    \textsuperscript{\rm 1} University of Maryland, Baltimore County, Information Systems Department\\ 

    \textsuperscript{\rm 2}The Hong Kong University of Science and Technology, CSE Department\\ 

    \{islam.rashidul,kkeya1,jfoulds,shimei\}@umbc.edu, \{zzengae, yqsong\}@cse.ust.hk
}

\begin{document}

\maketitle

\begin{abstract}
Recently, much attention has been paid to the societal impact of AI, especially concerns regarding its fairness.
A growing body of research has identified unfair AI systems and proposed methods to debias them, yet many challenges remain.  
Representation learning for Heterogeneous Information Networks (HINs), a fundamental building block used in complex network mining, has socially consequential applications such as automated career counseling, but there have been few attempts to ensure that it will not encode or amplify harmful biases, e.g. sexism in the job market.  
To address this gap, in this paper we propose a comprehensive set of de-biasing methods for fair HINs representation learning, including sampling-based, projection-based, and graph neural networks (GNNs)-based techniques.  
We systematically study the behavior of these algorithms, especially their capability in balancing the trade-off between fairness and prediction accuracy. 
We evaluate the performance of the proposed methods in an automated career counseling application where we mitigate gender bias in career recommendation. 
Based on the evaluation results on two datasets, we identify the most effective fair HINs representation learning techniques under different conditions. 
\end{abstract}

\section{Introduction}
Many real-world social and information networks such as social networks, bibliographic networks, and biological networks are heterogeneous in nature~\cite{LiuTwitterUncertain2016,ZhouCitationHIN2007,xiong2019DiseaseHeterogeneous}.
We can model such networks as Heterogeneous Information Networks (HINs) which contain diverse types of nodes and/or relationships. 
For example, we can model the Twitter social network as an HIN where the nodes are users or tweets, and the links are the follower/following relations between users and the authoring/retweeting relations between a user and a tweet.

To support large-scale heterogeneous information networks mining, much attention has recently been paid to representation learning for HINs where each node in a network is mapped to a dense vector in a low dimensional embedding space which preserves the relationships between nodes and important structural characteristics of the original networks~\cite{fu2017hin2vec,dong2017metapath2vec}. 
Due to its robustness, flexibility, and scalability, representation learning methods for HINs have been widely used to support diverse network mining tasks such as node classification~\cite{dong2017metapath2vec}, link prediction~\cite{wang2018shine}, community detection~\cite{cavallari2017learning}, and recommender systems~\cite{ShiHINEmbeddingRecommend2019}.

Despite its popularity, little attention has yet been paid to the understanding and mitigation of biases toward certain demographics in HINs representation, such as gender and racial biases. As demonstrated in a wide range of recent discoveries, machine learning systems trained with human-generated content (e.g., social media data) frequently inherit or even amplify human biases in the data ~\cite{angwin2016machine,dastin2018amazon,noble2018algorithms}. For example, word embedding models which inspired some of the early work on network embedding such as DeepWalk~\cite{perozzi2014deepwalk}, were shown to exhibit female/male gender stereotypes to a disturbing extent (e.g., ``man is to computer programmer as woman is to homemaker")~\cite{bolukbasi2016man}. To overcome this, there has been a concentrated recent effort in the natural language processing community~\cite{bolukbasi2016man,caliskan2017semantics,gonen2019lipstick} on understanding and mitigating the biases in word embeddings. In the network mining community, however, not much attention has been paid to the biases in HINs representation. Since HINs representation may encode harmful societal prejudice, it may cause unintended bias or unfairness in downstream applications. Therefore, it is important to ensure HINs representation is unbiased and applications incorporating these embeddings are fair and will not negatively impact vulnerable people and marginalized communities in our society.  

In this paper, we propose a range of fair HINs representation learning algorithms, including sampling-based, projection-based, and graph neural network (GNNs)-based methods to mitigate demographic bias in HINs representation. We systematically study the behavior of these algorithms, especially their capability in balancing the trade-off between prediction accuracy and fairness. 

To evaluate our algorithms,  we applied our fair HINs representation learning techniques to automated fair career recommendation. Career counseling plays an important role in many people's lives. Unbiased career advice based on an accurate assessment of one's interests, skills, and personality can help them make proper career choices. Good career choice in turn can boost their economic success, social standing, and quality of life. Biased career counseling however may restrict occupational opportunities and stunt the career development of disadvantaged populations (e.g., girls and minorities)~\cite{CareerCounselingWoman}. 

The main contributions of this work include:
\begin{itemize}
    \item To the best of our knowledge, this is the first systematic investigation on measuring and mitigating demographic bias in heterogeneous information networks representation.  
    Although prior work has studied mitigating bias in network embedding \cite{rahman2019fairwalk}, they focused on homogeneous networks instead of heterogeneous information networks.   
    \item We propose a comprehensive suite of de-biasing algorithms ranging from sampling-based, projection-based, to graph neural networks (GNNs)-based techniques to mitigate demographic biases in HINs representation.  
    \item We demonstrate the effectiveness of the proposed methods in mitigating gender bias in automated career counseling on two real-world datasets. Our results illuminate the prediction accuracy vs. fairness trade-off behavior of these algorithms, providing guidance to practitioners.
\end{itemize}
Our code is available at \url{https://github.com/HKUST-KnowComp/Fair_HIN}.

The rest of the paper is organized as follows. We start with a brief survey of related work, followed by a problem statement and the details of the proposed fair HINs representation learning algorithms. In the next section, we discuss the experiments designed to evaluate the effectiveness of the proposed methods under various conditions. We conclude the paper by summarizing the main findings and pointing out some future directions.

\section{Related Work}
In this section, we survey the research areas relevant to our paper. First, we summarize the main representation learning methods for heterogeneous information networks since they are the basis of our de-biasing algorithms. Next, we survey the general area of fair machine learning with a special focus on two most relevant subareas: fair word embeddings and fair homogeneous network embeddings. 

\subsection{HIN Representation Learning}
We categorize the typical HIN representation learning methods into two main types. Task-agnostic HIN representation learning methods focus on training general-purpose network embeddings that can be used to support different network mining tasks. In contrast, task-specific HIN representation learning methods learn representation via optimizing a particular downstream task. 

 
\subsubsection{Task-agnostic HIN Representation Learning} 
Task-agnostic HIN representation learning methods normally employ a meta-path guided sampling method to generate meta-paths. 
After obtaining meta-paths, typical word embedding algorithms such as word2vec \cite{mikolov2013efficient,mikolov2013distributed} are applied on the instantiated meta-path sequences to learn network embeddings.
A meta-path is a path consisting of a sequence of relations defined between different node types, which is either specified manually or derived from additional supervision \cite{fu2017hin2vec,dong2017metapath2vec,shi2018aspem}. 
For example, a meta-path \texttt{author} $\xrightarrow{\texttt{write}}$ \texttt{paper} $\xrightarrow{\texttt{written by}}$ \texttt{author} in a bibliographic network represents a co-authorship relation. \\

\subsubsection{Task-specific HIN Representation Learning}
GNNs-based methods are end-to-end supervised approaches to learn network embeddings. 
GNNs have the ability to aggregate local features, and to learn highly expressive representations. A comprehensive survey on GNNs \cite{wu2020comprehensive} shows that the majority of the current GNNs are designed for homogeneous networks. 
Recently, \cite{zhang2018deep} proposed a GNN method to explore heterogeneous networks. 
They translated an HIN into multiple homogeneous networks and applied a GNN to each homogeneous network and then combined them at the final layer.

\subsection {Fair Machine Learning: General Approaches}
There is a rising awareness that bias and fairness issues in machine learning (ML) algorithms can cause substantial societal harm \cite{angwin2016machine,buolamwini2018gender}. A concentrated effort in the machine learning community aims to address this problem. Existing methods on fair machine learning can be summarized into three general strategies. 

The first strategy employs fairness-aware pre-processing techniques to adapt the training data, including (a) modifying the values of sensitive attributes and class labels; (b) mitigating the dependencies between sensitive attributes and class labels by mapping the training data to another space \cite{DemoParityDwork2012,feldman2015certifying,rashid2019mitigating}, and (c) learning a fair representation of the training data that is independent of the protected attribute \cite{zemel2013learning,xie2017controllable}. 


The second strategy focuses on employing a fairness-guided optimization in model training. A fairness metric-based penalty is added as a constraint or a regularization term to the existing objective to enforce fairness  \cite{zafar2015fairness,zafar2017fairness,calders2010three,islam2021debiasing}. For example, \cite{zafar2015fairness} minimizes the covariance between the sensitive attributes and the (signed) distance between feature vectors and the decision boundary of a classifier.

The third strategy modifies posteriors to satisfy fairness constraints.  For example, \cite{hardt2016equality} selected a threshold such that the true positive rates of different groups are equal.

Furthermore, there are existing works using post-processing techniques to correct the bias depending on the protected attribute. For example, \cite{d2017conscientious} proposed to audit fairness of the model on a held-out set to mitigate bias by altering some predicted outputs.

\subsection{Fair Word Embeddings}
Word embedding models \cite{mikolov2013distributed,mikolov2013efficient} learn a mapping of each word in the vocabulary to a vector in an embedding space to encode semantic meaning and syntactic structures of natural languages. 
These models are typically trained using a neural network-based representation learning algorithm on word co-occurrence data computed from massive text corpora. Since text data may contain societal stereotypes such as racism or sexism, word embeddings also typically inherit or amplify biases present in the data \cite{bolukbasi2016man,caliskan2017semantics,papakyriakopoulos2020bias}.

The most popular method of de-biasing word embeddings is to project each word embedding orthogonally to the bias direction \cite{bolukbasi2016man} followed by a crowd-sourcing based correction. 
Instead of relying on the crowd to find gendered words,
alternatively, the demographic bias direction can simply be computed using gendered person names \cite{dev2019attenuating}. 

\subsection{Fair Homogeneous Network Embedding}
Fair network embedding is a relatively new area that is now beginning to be addressed. 
In an approach related to ours, \cite{rahman2019fairwalk} modified random walks to learn fairness-aware embeddings. At each step, the algorithm partitions neighbors into different groups based on the values of the sensitive attribute. 
The system tries to give each group the same probability of being selected regardless of its size. 
The method is applicable only to homogeneous networks, while our work addresses heterogeneous information networks.  
\cite{bose2019compositional} proposed an adversarial framework to learn fair graph embeddings. 
They trained an encoder to optimize a downstream task (i.e., link prediction task in knowledge graphs) as well as to filter out information about sensitive attributes.
They also trained a discriminator to predict the sensitive attribute from the graph embeddings. 
This algorithm can be applied to both homogeneous and heterogeneous networks. 
But they did not utilize the property of heterogeneous information networks to learn graph embeddings and to enforce fairness. 

\section{Problem Statement}
\begin{figure}
    \center
    \includegraphics[width=0.3 \textwidth]{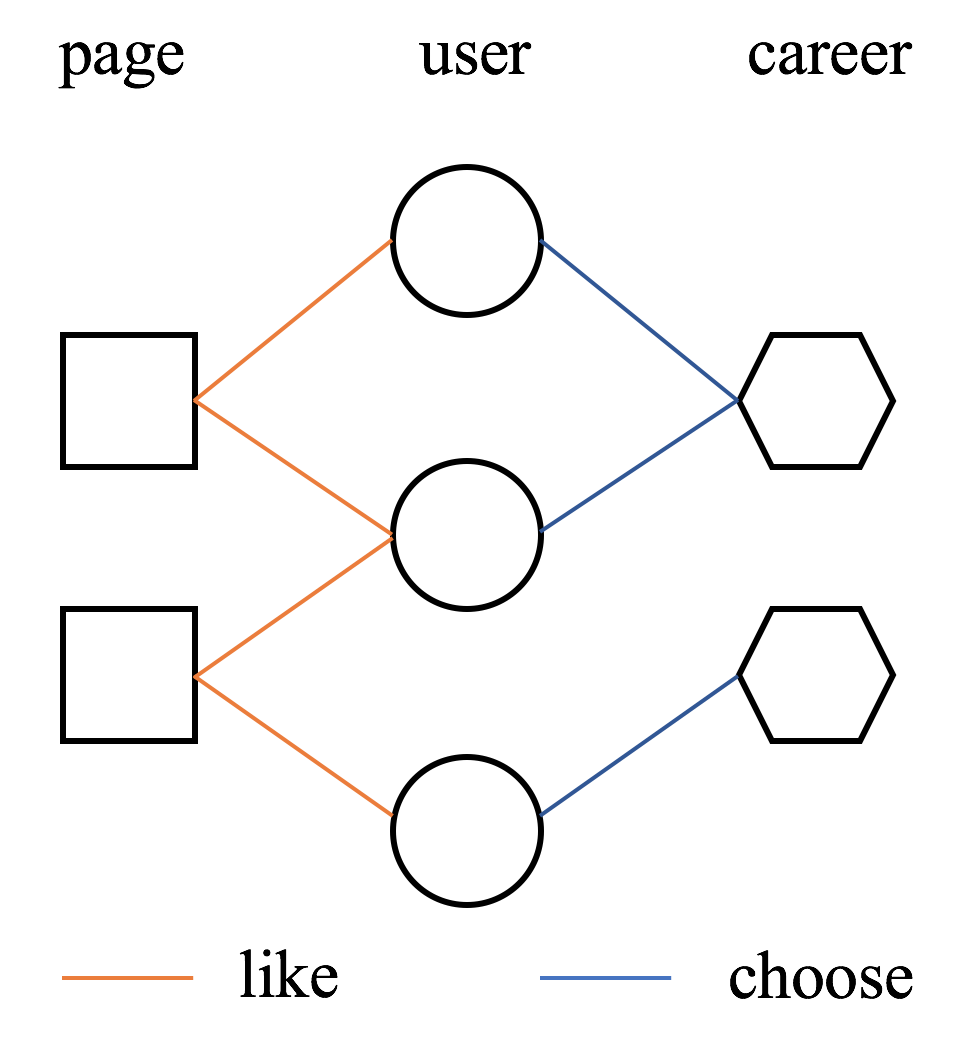}
    \caption{An illustrative example of a Facebook career network. Different shapes (colors) indicate different types of nodes (relations). A square represents a page. A circle represents a user. A hexagon represents a career. Orange line represents ``like'' relation. Blue line represents ``choose'' relation. }\label{fig:hin}
\end{figure}

We first describe our problem setting.    
We assume that our dataset is a heterogeneous information network $G=(V,E,T,R)$ with $|T| > 1$, where $V$ denotes a set of nodes, $E$ denotes a set of edges, $T$ denotes a set of node types (e.g., user, career, and item), and $R$ denotes a set of edge relations (e.g., a user likes a Facebook page, a user rates a movie, and a user chooses a career).  We also assume a binary protected attribute $a$, which pertains to at least one type of nodes(e.g. gender for the users). An illustrative example of a heterogeneous information network (Facebook career network) is shown in Figure \ref{fig:hin}.  

We focus on fairness in the general task of link prediction in heterogeneous information networks, which we illustrate with an application to career counseling.
Our goal is to learn node embeddings $e_v$ for all nodes $v \in V$ such that their use for link prediction on held-out edges leads to accurate performance, and the system behaves in a fair manner with regard to the protected demographics $a$. 
Fairness is a complex socio-technical issue. So far many fairness definitions have been proposed in the AI community. Here we adopt two of the most widely used AI fairness definitions and consider their applications to link prediction or node classification in heterogeneous information networks, e.g. for career counseling. 
Career counseling aims to automatically provide a career choice for users based on their interests (e.g., clicking ``like'' on Facebook pages). \\

\noindent\textbf{Demographic Parity}
Demographic Parity is one of the most well-known criteria for fairness in machine learning classifiers~\cite{DemoParityDwork2012}. It is defined as:
$$ P(\hat{y}=k|a=0) = P(\hat{y}=k|a=1) \; \forall k ,$$
where $\hat{y}$ is a predicted label, $y$ is the ground truth class label, $a$ is the protected attribute such as gender (here we only consider binary protected attributes), and $k$ represents the possible values of the class label $y$. The extent to which the demographic parity criterion is violated is measured by a distance metric between the two conditional distributions, typically chosen to be the total variation distance (see Eq. (\ref{eq:diff_dp})). 

In the case of link prediction in heterogeneous information networks, we adapt the definition by letting $P(\hat{y}=k,a=i)$ be the empirical probability of predicting a node as class label $k$ (e.g. a career such as computer science) and the node is of protected type $i$ (e.g. a female user).  
Based on this definition, in career counseling, to achieve perfect demographic parity, the probability of recommending each career (e.g. computer science) given male users should be the same as that given female users. \\

\noindent\textbf{Equal Opportunity}
Another popular fairness definition is equal opportunity \cite{hardt2016equality}. It is defined as:
  $$ P(\hat{y}=k|a=0,y=k) = P(\hat{y}=k|a=1,y=k) \; \forall k \mbox{ .}$$
We adapt this to link prediction in heterogeneous information networks similar to demographic parity.  
we adapt the definition by letting $P(\hat{y}=k,a=i,y=k)$ be the empirical probability of predicting a node as class label $k$ (e.g. a career such as computer science) and the ground truth is class label $k$ and the node is of protected type $i$ (e.g. a female user).  Based on this definition, in career counseling, to achieve perfect equal opportunity, for a subgroup of people who indeed have chosen a particular career (e.g., computer science),  the probability of the system to recommend this career (e.g., computer science) given male users should be the same as that given female users.

\section{Methodology}

In this section, we present three classes of fair HIN representation learning methods: (1) \textit{fairness-aware sampling} which extends existing sampling methods for HIN embedding to mitigate bias, (2) \textit{embedding projection} which employs a vector-space projection operation to reduce biases in the embeddings, and (3) \textit{graph neural networks (GNNs) with a fairness-aware loss}, which trains a GNN to optimize both prediction accuracy and fairness.  

To illustrate these algorithms, we use fair career counseling as an example application. We formulate fair career counseling as a fair link prediction in HINs. Here, the nodes in the HIN includes  a \textit{user} (e.g., a Facebook user or a movie reviewer), an \textit{item} (e.g., a  Facebook page or a movie) and a \textit{career} (e.g., the \textit{career concentration} or \textit{occupation} declared by a user). A user node is linked to an item node via a ``like'' link, indicating a user's preference for the item. A user node can also be linked to a ``career'' node via a ``choose'' link.  The goal of automated career recommendation is to predict whether there should be a link between a user and a career based on the interests/likes of a user. 
In the following, we propose a range of methods for fair HIN representation learning.

\subsection{Fairness-aware Sampling}


To learn embeddings in a HIN, we often use meta-paths to guide a random walker to generate traversal paths.   
A meta-path is a path consisting of a sequence of relations defined between different node types.
For example, \texttt{user} $\xrightarrow{\texttt{choose}}$ \texttt{career} $\xrightarrow{\texttt{chosen by}}$ \texttt{user} represents two users choosing the same career. After obtaining meta-paths, we can use a meta-path based HIN representation learning algorithm such as metapath2vec (M2V) to learn the node embeddings.
M2V incorporates the heterogeneous network structures into the skip-gram model \cite{mikolov2013distributed,mikolov2013efficient}.  
Specifically, the objective function of M2V is defined as follows,
\begin{equation} \label{eq:obj_j1}
\argmax_{\theta} \sum_{v \in V} \sum_{t \in T} \sum_{c_t \in N_t(v)} \log p(c_t|v;\theta),
\end{equation}
where $N_t(v)$ denotes node $v$'s neighbors which are of type $t$, and $p(c_t|v;\theta)$ is commonly defined as a softmax function. 
The objective function of M2V is similar to that of the skip-gram model except that M2V is sensitive to the type of node.

After learning user embeddings and career embeddings, we then train a multilayer perceptron (MLP) to predict the career, where the input is a user embedding and the career embeddings learned by M2V.  

\subsubsection{Traditional Meta-path Generation}

A meta-path $\mathcal{P}$ is a path defined on the graph $G = (V, E, T, R)$, and is denoted in the form of $V_{1} \xrightarrow{R_{1}} V_{2} \xrightarrow{R_{2}} \cdots \xrightarrow{R_{l}} V_{l+1}$. 
This meta-path defines a composite relation $R = R_1 \circ R_2 \circ \cdots \circ R_l$ between type $1$ and type $l+1$, where $\circ$ denotes the composition operator on relations, and $V_t$ denotes all  nodes with type $t$.

Here we show how to use meta-paths to guide heterogeneous random walkers. 
The transition probability at step $i$ is defined as follows,
\begin{equation}
	\label{eq:normal_meta}
	P(v^{i+1}|v^{i}_{t}) =  
    \begin{cases}
		\frac{1}{|N_{t+1}(v^{i}_{t})|}, & (v^{i+1},v^{i}_{t}) \in E, \phi(\cdot) = t+1 \\
        0, & (v^{i+1},v^{i}_{t}) \in E, \phi(\cdot) \neq t+1 \\
		0, & (v^{i+1},v^{i}_{t}) \not\in E
	\end{cases}
    ,
\end{equation}
where $v^{i}_{t} \in V_t$, and $N_{t+1}(v^{i}_{t})$ denotes $v^{i}_{t}$'s neighbors which are of type $t+1$, and $\phi(\cdot)$ is a function which returns the type of the node $v^{i+1}$. The transition probability function shows that the random walker only goes to a certain type of nodes and the type of next step is guided by meta-paths. 

We manually define two meta-paths for our career counseling application. 
\texttt{career} $\xrightarrow{\texttt{chosen by}}$ \texttt{user} $\xrightarrow{\texttt{like}}$ \texttt{item} $\xrightarrow{\texttt{liked by}}$ \texttt{user} $\xrightarrow{\texttt{choose}}$ \texttt{career} represents two careers chosen by two different users who like the same item (e.g., a Facebook page, or a movie). 
\texttt{user} $\xrightarrow{\texttt{like}}$ \texttt{item} $\xrightarrow{\texttt{liked by}}$ \texttt{user} represents two users like the same item. 

\subsubsection{Fair Meta-path Generation}

Assume $a$ is a protected binary attribute and $i$ is a value of $a$, $g_i$ represents a group of users satisfying $a=i$. If $a$ is a binary attribute, we define an advantaged group if it contains a larger number of training examples than the other group (the disadvantaged group). Fair meta-path generation aims to up-sample the disadvantaged group with higher probability while down-sample the advantaged group. The sampling probability is inversely proportional to the number of users in each group. As the number of users in the disadvantaged group is less than that in the advantaged group, the disadvantaged group has a higher probability to be sampled. As a result,  the system is less likely to neglect the disadvantaged group. 

Next, we present the details of the algorithm. Fair meta-path sampling occurs at the step where the current node type is \textit{career} ($C$) and the next node type is \textit{user} ($U$). Assume $a$ is a protected attribute of a user (e.g., gender or race).  To simplify our explanation, we assume $a$ is binary. Based on the values of $a$, we can cluster all the users into different groups (e.g., $g_0$ and $g_1$).  The unnormalized transition probability at step $i$ where $t=C$ and $t+1=U$ is defined as follows:
\begin{equation}
	\label{eq:career_meta}
	P(v^{i+1}|v^{i}_{C}) =  
    \begin{cases}
        \frac{r}{|N(v^{i}_{C},U,g_0)|}, & cond, \pi(v^{i+1}) = g_0 \\
        \frac{r}{|N(v^{i}_{C},U,g_1)|}, & cond, \pi(v^{i+1}) = g_1 \\
        0, & (v^{i+1},v^{i}_{C}) \in E, \phi(\cdot) \neq U \\
		0, & (v^{i+1},v^{i}_{C}) \not\in E
	\end{cases},
\end{equation}
where $\phi(\cdot)$ is a function which returns the type of node $v^{i+1}$, and $\pi(\cdot)$ is a function which returns the value of the protected attribute(e.g., the gender of a user) of node $v^{i+1}$; $cond$ is the condition where $(v^{i+1},v^{i}_{C}) \in E, \phi(\cdot) = U$, 
which means the current node is a career node and the next node $v^{i+1}$ is a user node, and they are connected; $N(v^{i}_{C},U,g_0)$ denotes the neighbors of $v^{i}_{C}$ which are of type $U$ and belong to group $g_0$, and $N(v^{i}_{C},U,g_1)$ denotes the neighbors of $v^{i}_{C}$ which are of type $U$ and belong to group $g_1$; and $r$ is a hyper-parameter that can determine to what extend the random walker selects the disadvantaged group in order to correct the bias. Note that if $|N(v^{i}_{C},U,g_0)| = 0$ or $|N(v^{i}_{C},U,g_1)| = 0$, then the corresponding unnormalized probability should be zero. 
The transition probability in the second and third conditions remains the same as that in Eq. (\ref{eq:normal_meta}).

\subsection{Embedding Projection}
The second debiasing method is inspired by recent work on attenuating bias in word vectors \cite{dev2019attenuating}. 
It is often used as a post-processing step.  We adapt this method to de-bias HINs embeddings. 
Basically, after obtaining HINs embeddings, we can reduce their unfairness by eliminating the effect of a protected attribute (e.g., gender) in learned user embeddings via vector projection. The main difference between our method and the projection method used in \cite{dev2019attenuating} is the bias direction in the embedding space. 

\cite{dev2019attenuating} computed a ``bias direction'' for word embeddings based on the difference in the average embeddings of male and female names.  In our case, 
let $v_{g_i}$ be the direction of group $g_i$. We compute $v_{g_i}$ by averaging all the embeddings of the users in group $g_i$,
\begin{equation}
    v_{g_i} = \frac{e_{u_1}+e_{u_2}+\cdots+e_{u_{n_i}}}{\Vert e_{u_1} + e_{u_2} + \cdots + e_{u_{n_i}} \Vert},
\end{equation}
where $e_{u_1}$, $e_{u_2}$, $e_{u_{n_{i}}}$ are the user embeddings in $g_i$, and there are $n_i$ users in group $g_i$. 
Assuming the protected attribute is binary, we can compute the bias direction $v_b$ using
\begin{equation}\label{eq:bias_direction_1}
    v_b = \frac{v_{g_0} - v_{g_1}}{\Vert v_{g_0} - v_{g_1} \Vert}  \mbox{ .}
\end{equation}
 




To reduce bias in user embeddings, we project each user's vector $e_u$ orthogonally to the bias direction $v_b$ as follows, 
\begin{equation}
    e_u' = e_u - <e_u,v_b>v_b,
\end{equation}
where $v_b$ is the bias direction, $<\cdot,\cdot>$ is  the inner product operation, and $e_u'$ is the resulting de-biased user embedding. 

\subsection{Fairness-aware Graph Neural Networks (GNNs)-based Representation Learning}
In this section, we explore task-specific representation learning methods, i.e., graph neural networks (GNNs). 
GNNs learn the node representation via optimizing a downstream task (e.g., node classification) in a supervised manner. 
In the career counseling application, task-agnostic representation learning methods (e.g., M2V) consider that a career is a node and users nodes are linked to career nodes. 
While GNNs methods consider that a career is the label of a user node, so there are only two types of nodes (users and items) in the input for GNNs models.   
Task-agnostic representation learning methods consider career recommendation as a link prediction task, GNNs methods formulate it as the node classification task. 
%
To reduce bias, we directly incorporate a fairness-aware loss in addition to an accuracy-based objective. 
In the following, we describe our method to make GNNs representation fair. 
 
\subsubsection{Graph Neural Networks}
Graph Neural Networks (GNNs)~\cite{kipf2016semi,william2017representation,peter2018graph} are powerful tools for graph mining. 
It has gained increasing popularity in various applications, including social networks, knowledge graphs, recommender systems, and biomedical research.  

Let $G = (V, E)$ denote a graph with node features $X_v$ for $v \in V$. 
Given a set of nodes $\{v_1, ..., v_n\}$ and their labels $\{y_1, ..., y_n\}$, the task of GNNs is to learn a representation vector $h_v$ for node $v$ that helps to predict the label of the node $v$. The prediction is denoted as $\hat{y}$. 
In GNNs model, the representation of node $v$ is iteratively updated by aggregating the representations of $v$’s neighboring nodes. After $l$ iterations of aggregation, $v$’s representation captures the structural characteristics of its $l$-hop network neighborhood. 
Formally, the representation of node $v$ in $l$-th layer in a GNN is formulated as follows,
\begin{equation}
h^l_v = Combine \big(h^{l-1}_v,Agg(h^{l-1}_v,h^{l-1}_u)\big),
\end{equation}
where $h^{l}_{v}$ is the representation of node $v$ at the $l$-th iteration/layer, and $u$ are neighboring nodes of $v$, and $Combine$ is a function to combine the information of node $v$ and its neighboring nodes in $l-1$ th layer, and $Agg$ is an aggregation operation which aggregates neighbors of $v$. 
Commonly used aggregator operations are mean aggregator, LSTM aggregator, and pooling aggregator. We initialize $h^0_v = X_v$.
The accuracy loss function of GNNs is:
\begin{equation}
\LM_{acc} = - \sum_{i}^{N} { \log P(\hat{y_i}=y_i)} \mbox{ .}
\label{eq:gnn_acc}
\end{equation}

\subsubsection{Fairness-aware Loss}
There are many fairness definitions. 
Here we only consider demographic parity \cite{DemoParityDwork2012} and equal opportunity \cite{hardt2016equality}. 

We introduce some notations first. $N^{i}$ denotes the number of users in protected group $g_i$. 
Here, we only consider binary protected attributes, hence, $i \in \{0,1\}$.
$N^{i}_{k}$ denotes the number of samples which are predicted to be $k$ and are in the protected group $g_i$.
$N^{i}_{k,k}$ denotes the number of samples which are predicted correctly and are in protected group $g_i$. 
The computation of empirical demographic parity and equal opportunity is:
\begin{equation}\label{eq:diff_dp}
\begin{split}
\textit{diff}_{dp} & = \sum_{k} { \bigg|\frac{N^{0}_{k}}{N^{0}} - \frac{N^{1}_{k}}{N^{1}}\bigg|}
\end{split},
\end{equation}

\begin{equation}\label{eq:diff_eo}
\begin{split}
\textit{diff}_{eo} & = \sum_{k}{ \bigg|\frac{N^{0}_{k,k}}{N^{0}_{k}} - \frac{N^{1}_{k,k}}{N^{1}_{k}}\bigg|}
\end{split}.
\end{equation}

The above computation is based on counting, which cannot be used as a loss function, because the gradient cannot be back propagated. Hence we use the probabilistic prediction $p(\hat{y}|x)$ of the model to replace the hard count. $x$ is a sample in the dataset. Specifically, in career counseling application, $x$ is a user. Hence, we define the fairness-aware loss for GNN as follows,
\begin{equation} \label{eq:dp_loss}
\begin{split}
\LM_{dp} & = \sum_{k} \bigg( \frac{\sum_{x \in D:a=0}{P(\hat{y}=k|x)}} {N^{0}} \\
& \quad \quad \quad - \frac{\sum_{x \in D:a=1}{P(\hat{y}=k|x)}} {N^{1}} \bigg)^2,
\end{split}
\end{equation}

\begin{equation} \label{eq:eo_loss}
\begin{split}
\LM_{eo} & = \sum_{k} \bigg( \frac{\sum_{x \in D:a=0,y=k}{P(\hat{y}=k|x)}}  {N^{0}_{k}} \\
& \quad \quad \quad - \frac{\sum_{x \in D:a=1,y=k}{P(\hat{y}=k|x)}}{N^{1}_{k}} \bigg)^2,
\end{split}
\end{equation}
where $x$ denotes a sample; $D:a=i$ is the set of samples where their values of the attribute are $i$; $D:a=i,y=k$ is the set of samples where their values of the attribute is $i$ and ground truth is $k$. 
We compute two final loss functions in our GNN models. 
The first one is a demographic parity based fairness-aware loss function:
\begin{equation}\label{eq:com_dp}
    \LM_{acc} + \lambda_{dp} \cdot \LM_{dp},
\end{equation}
where $\lambda_{dp}$ is a trade-off hyper-parameter.

The second one is an equal opportunity based fairness-aware loss function:
\begin{equation}\label{eq:com_eo}
    \LM_{acc} + \lambda_{eo} \cdot \LM_{dp},
\end{equation}
where $\lambda_{eo}$ is a trade-off hyper-parameter. 
\section{Experiments}
In this section, we study the behavior of the proposed de-biasing methods for HINs representation and their applications in automated career counseling. We tested the their performance on two datasets: a Facebook dataset and a MovieLens dataset.

\subsection{Datasets}
\begin{table}[th!]
    \centering
    \begin{tabular}{llll}
        \toprule
        Dataset & Facebook & MovieLens \\
        \midrule
        
        \# careers & 48 & 14 \\
        \# items (FB pages or movies) & 99,756 & 3,677 \\
        \# users & 7,069 & 4,920 \\
        \# male users & 2,721 & 3,558 \\
        \# female users & 4,332 & 1,362 \\
        avg users per career & 62.81 & 93.47 \\
        avg users per item & 14.04 & 222.40 \\
        avg items per user & 198.15 & 166.21 \\
        \bottomrule
    \end{tabular}
    \label{table:dataset_stat}
    \caption{Statistics of Facebook and MovieLens datesets.}
\end{table}
The Facebook dataset used in the study was collected as a part of the myPersonality project~\cite{kosinski2015facebook}. The data was collected with an explicit opt-in consent for reuse
for research purposes. To protect privacy, the data were anonymized. 

The Facebook dataset contains rich information about a Facebook user such as his/her demographics (e.g., gender), the Facebook pages he/she likes, and his/her declared career concentrations (e.g., music, computer science, psychology). Since Facebook ``likes'' contain rich information about a person's interests and preferences of a wide range of items/topics (e.g., books, music, celebrities, brands, and TV shows), they can be used to suggest possible career concentrations for a new user.  

As the myPersonality dataset is no longer publicly available, to facilitate research reproducibility, we employed a second dataset, the movieLens dataset, which is publically available. It contains similar information to the Facebook dataset such as the gender of a movie reviewer, the movies he/she likes, and his/her occupation. 
 

Table 1  shows the statistics of each dataset after we clean the data. The Facebook network created from the myPersonality dataset  consists of $7,069$ user nodes, $99,756$ item nodes (a.k.a FB page), and $48$ career nodes. If a user likes a page, there is an edge between the user and the page. If a user declared a career concentration on Facebook, there is an edge between the user and the career.


The MovieLens-1M dataset originally contains $6,040$ users, $3,900$ movies, and $19$ careers. After we remove some careers that our system should not recommend such as ``retired,'' ``unemployed,'' and ``K-12 student,'' the resulting MovieLens network consists of $4,920$ user nodes, $3,677$ movie nodes, and $14$ career nodes. Similarly, there is an edge between the user and the movie if a user rated a movie. If a user has declared an occupation, there is an edge between the user and the occupation.  


In both datasets, each user only has one career. 
In the Facebook dataset, we filtered out users for which binary gender information was not provided by the user. 
All users in MovieLens had binary gender information provided. 
We were unable to study our method's impact on non-binary identifying users as non-binary gender information was not provided. 
 
Since our datasets are not very large, to fully utilize our ground truth data,  we employed nested cross validation for model training, testing, and hyper-parameter tuning. Specifically, we first split the whole dataset into two parts, where 40\% of the data were used for embeddings training, and the remaining 60\% were used for career prediction. For the data reserved for career prediction, we split it further into three folds, with one of the folds for testing, and the other two folds for training and validation. We further split the training and validation data into four folds with one of the folds as a validation set, and the other three folds for training. The proportion of embeddings training, career prediction training, career prediction validation, and career prediction test was 4:3:1:2. 
 
For GNN based methods, to facilitate result comparison, the test sets were the same as those used in the sampling-based methods. Since GNNs do not need any data to train embeddings, the data for embeddings training (i.e., 40\% of the whole data) was added into the career prediction training data, so the proportion of training, validation, and test for career prediction was 7:1:2. 

\subsection{Evaluation Metrics}
We use one prediction accuracy measure and two fairness measures to assess the performance of all the methods. \\

\noindent\textbf{Accuracy Measure}: we use Mean Reciprocal Rank (MRR) as the measure of prediction accuracy. It is a statistical measure for evaluating an ordered list of items. MRR is computed as the mean of the multiplicative inverse of the rank of the correct answers. MRR is frequently used in information retrieval and recommender systems to assess the system output.  To compute MRR, let $rank_i$ denote the rank of the ground truth (i.e, a user's career choice), and $N$ denote the number of samples. Then $MRR=\frac{1}{N}\sum_{i}^{N}{1/rank_i}$.\\

\noindent\textbf{Fairness Measures}: we use two widely used fairness measures in our evaluation: Demographic Parity and Equal Opportunity, which are defined in Eq. (\ref{eq:diff_dp}) and Eq. (\ref{eq:diff_eo}).

\subsection{Compared Methods}
We have implemented a total of eight different methods. 
Among them, one is traditional HINs embedding methods without gender de-biasing (M2V), and one is graph neural network methods without gender de-biasing (GNN). The rest are fairness-aware methods. Among the seven fairness-aware methods, one is a naive baseline (called balance-data). 
One is the adversarial baseline proposed by \cite{bose2019compositional}. 
The rest of the methods employ a single or a combination of the de-biasing methods we proposed. The details of each method are described below.\\

\noindent \textbf{\textit{balance-data}}: For each career, we randomly remove users from the advantaged group so that we have an equal number of male and female users in the embedding training data. In this way,  we deliberately create a balanced dataset to remove gender bias. Note that only the embeddings training set is balanced. The career prediction training/validation/test sets remain the same to facilitate fair comparison. After creating a balanced dataset, we use the metapath2vec algorithm \cite{dong2017metapath2vec} to train embeddings. \\

\noindent \textbf{\textit{adversarial}:} \cite{bose2019compositional} proposed an adversarial framework to learn fair graph embeddings. 
They trained an encoder to learn graph embeddings as well as to filter out information about sensitive attributes.
They also trained a discriminator to predict the sensitive attribute from the graph embeddings. 
The encoder we used in the experiment is the GraphSaint model \cite{zeng2019graphsaint}. 
The discriminator we used in the experiment is an MLP which has the same architecture in \cite{bose2019compositional}. 
Similar to GNNs-based method, we formulate the career prediction task to a node classification task. \\

\noindent \textbf{\textit{M2V}}: we use the traditional meta-path generation algorithm in Eq. (\ref{eq:normal_meta}) to generate meta-paths, and learn user embeddings and career embeddings using metapath2vec algorithm \cite{dong2017metapath2vec}. \\

\noindent \textbf{\textit{M2V + fair sampling}}: we use fair meta-path generation algorithm shown in Eq. (\ref{eq:career_meta}) to generate meta-paths, and the rest of the processes are the same as M2V.\\

\noindent \textbf{\textit{projection}}: we calculate the bias direction using Eq. (\ref{eq:bias_direction_1}) 
and project all the user embeddings to the direction that is orthogonality to the bias direction. 
All methods which learn embeddings explicitly can use the projection method as a post-processing step. \\

\noindent \textbf{\textit{GNN}}: we use the GraphSaint algorithm \cite{zeng2019graphsaint}, a state-of-the-art GNN algorithm to predict career. We have also tried GraphSAGE \cite{hamilton2017inductive}. However, the results were much worse than GraphSaint. So in this paper, we only report the results using GraphSaint. 
We formulate the career prediction task to a node classification task, where the career choice of a user is the label of the user node. \\

\noindent \textbf{\textit{GNN-demographic-parity}}: we use Eq. (\ref{eq:com_dp}) which combines the demographic-parity based fairness-aware loss and the accuracy loss to train the GraphSaint model \cite{zeng2019graphsaint}.\\

\noindent \textbf{\textit{GNN-equal-opportunity}}: we use Eq. (\ref{eq:com_eo}) which combines the equal-opportunity based fairness-aware loss with the accuracy loss to train the GraphSaint model \cite{zeng2019graphsaint}.

For the sampling-based based methods above, after we learn the embeddings,  we train a multilayer perceptron (MLP) to predict the career. 
The inputs are the user embeddings and the career embeddings learned by sampling-based methods. 
Since the output of the softmax layer in MLP is a probability distribution, we rank all career candidates based on their corresponding probabilities, and use mean reciprocal rank as the measure of prediction accuracy. 

For all sampling-based methods, we used the same hyper-parameters listed below.
The dimension of embeddings was $128$; the size of negative samples was $5$; the context window size was $5$.  
For all GNNs-based methods and the encoder in the adversarial method, we used two convolutional layers. The dimension of the hidden layer is $128$.  The features of nodes are generated using the following procedure. 
Each Facebook page or movie is associated with a title or description. 
We average the word embeddings of all the words in the title or description as its features. 
For each user, we average the features of all the linked items (e.g., FB page or movie) as his/her features. For each career, we average the features of all linked users as its features. The dimension of the features was $50$.

\begin{figure*}[t]
	\centering
    \subfigure[MRR vs demographic parity on Facebook]{\label{fig:fb_dp}
		\includegraphics[width=0.48\textwidth]{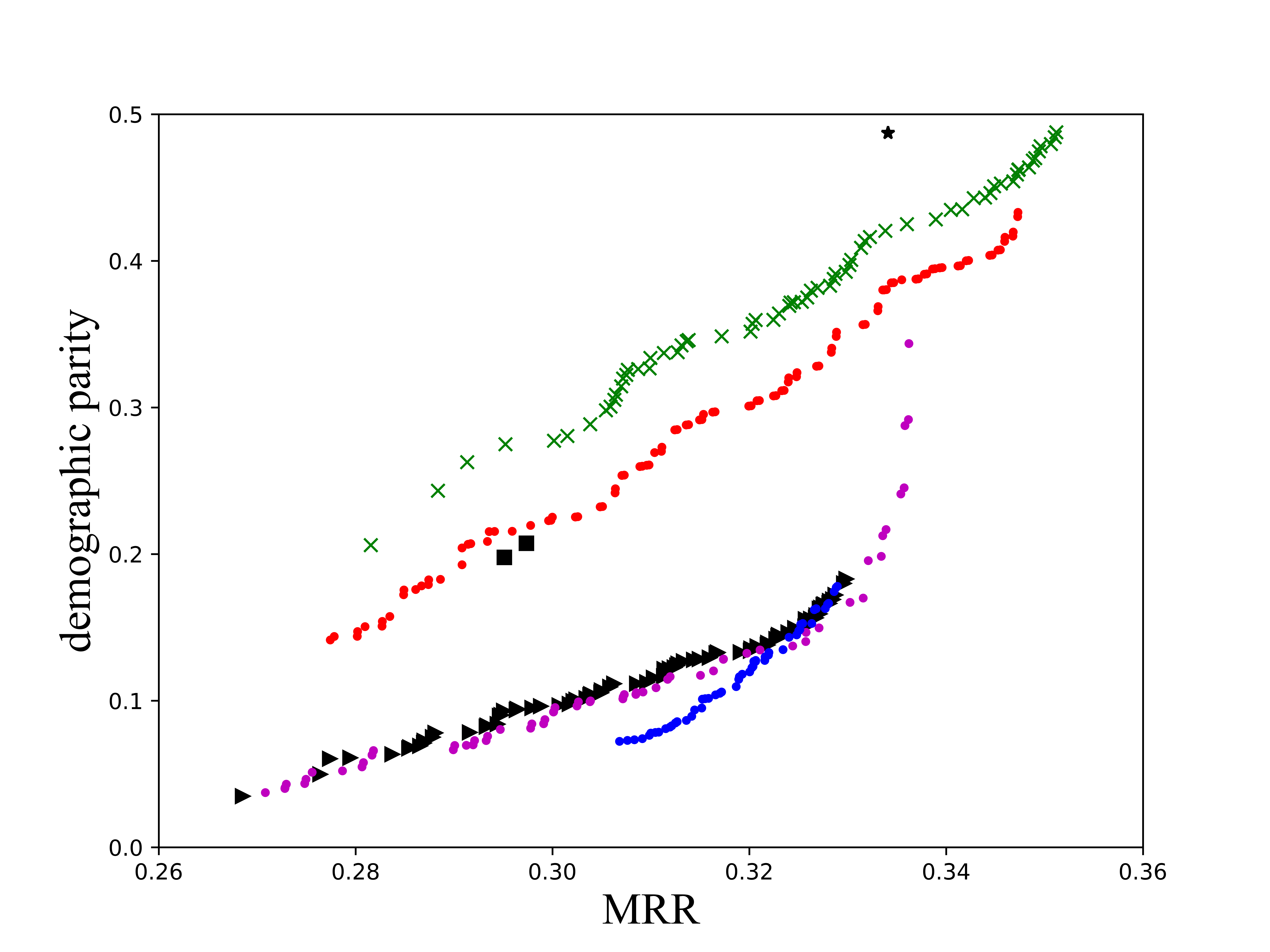}
	}
	\subfigure[MRR vs equal opportunity on Facebook]{\label{fig:fb_eo}
		\includegraphics[width=0.48\textwidth]{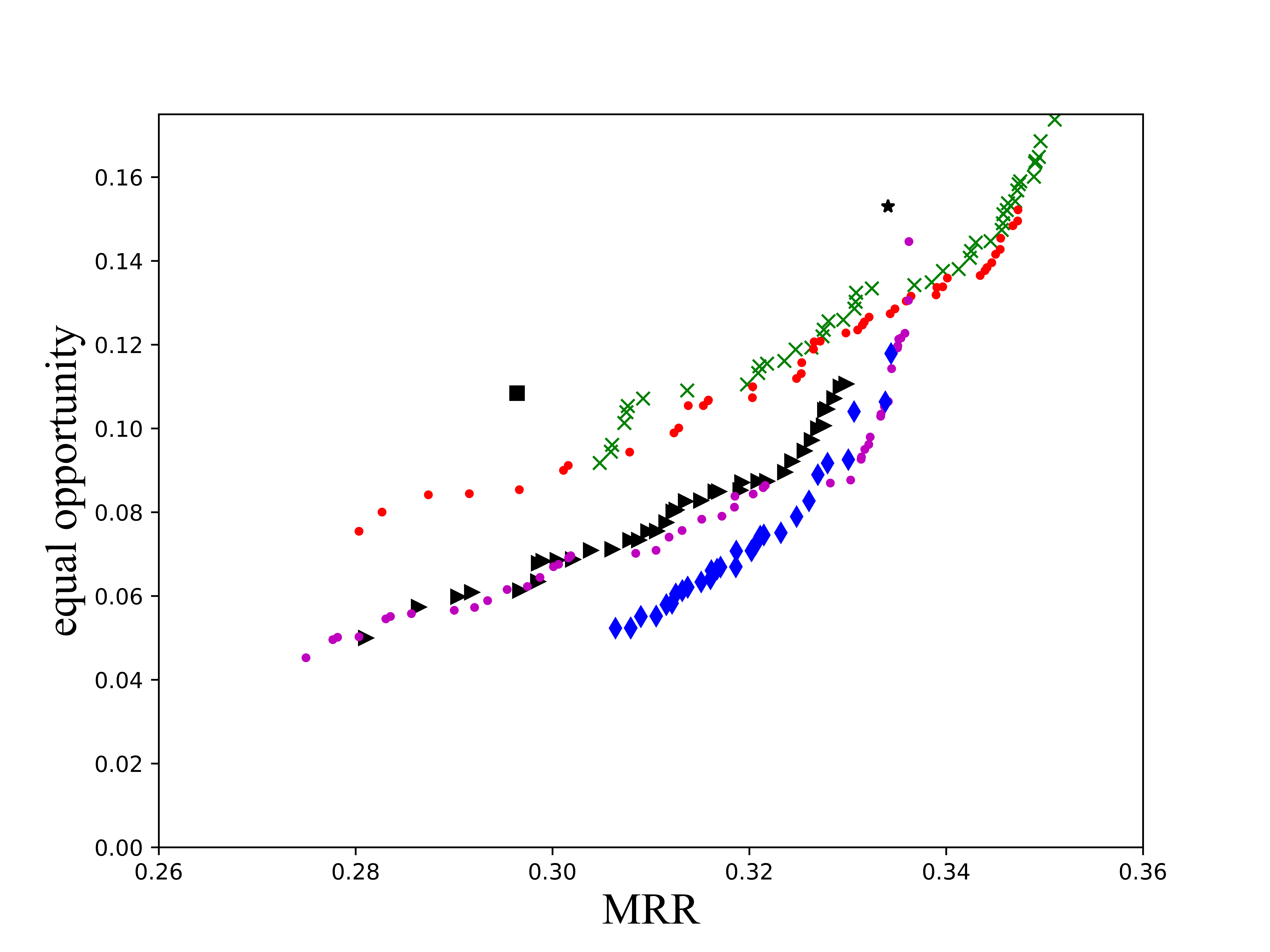}
	}
	\\
	\vspace{-0.15in}
	\subfigure[MRR vs demographic parity on MovieLens]{\label{fig:ml_dp}
		\includegraphics[width=0.48\textwidth]{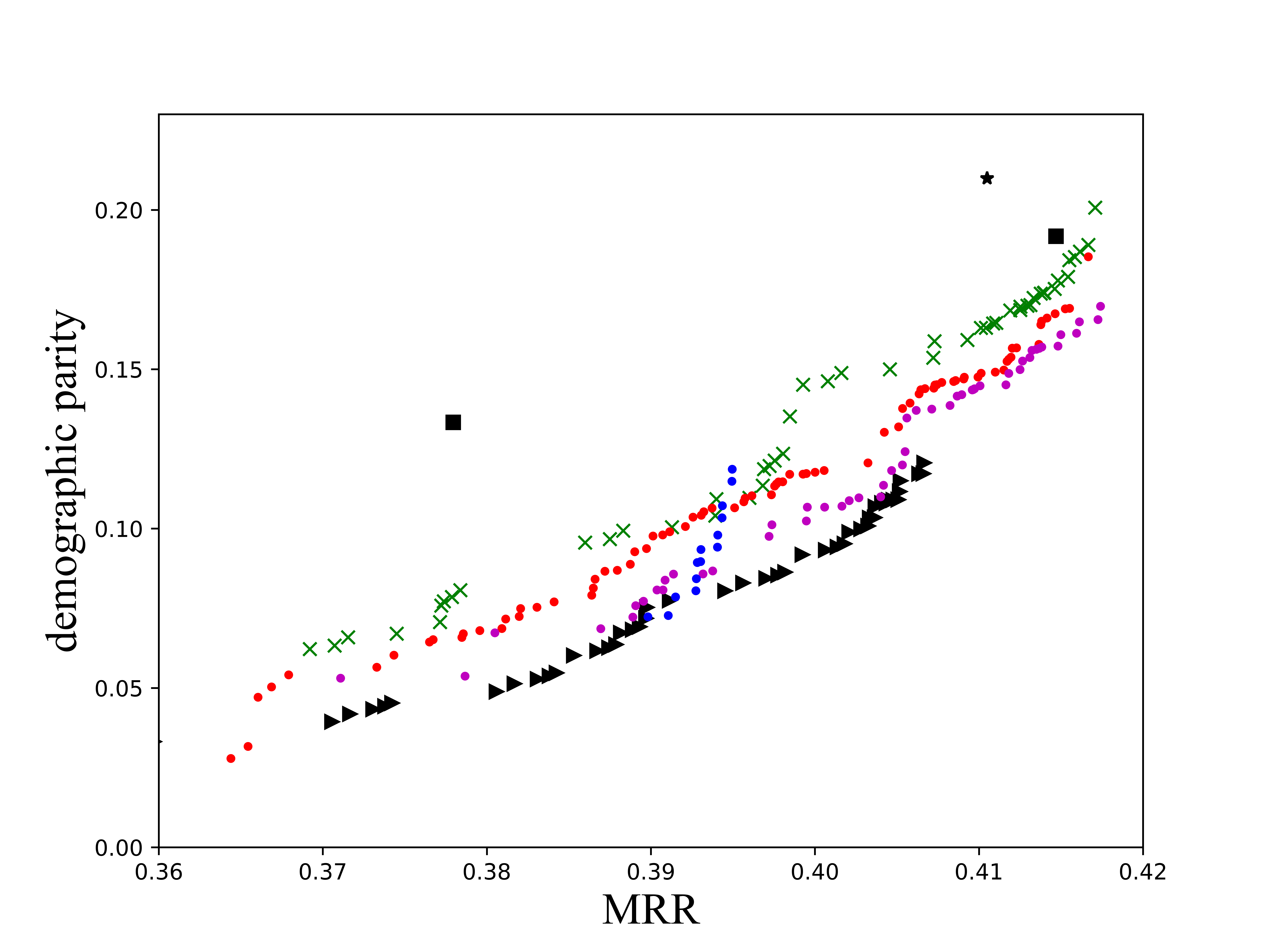}
	}
	\subfigure[MRR vs equal opportunity on MovieLens]{\label{fig:ml_eo}
		\includegraphics[width=0.48\textwidth]{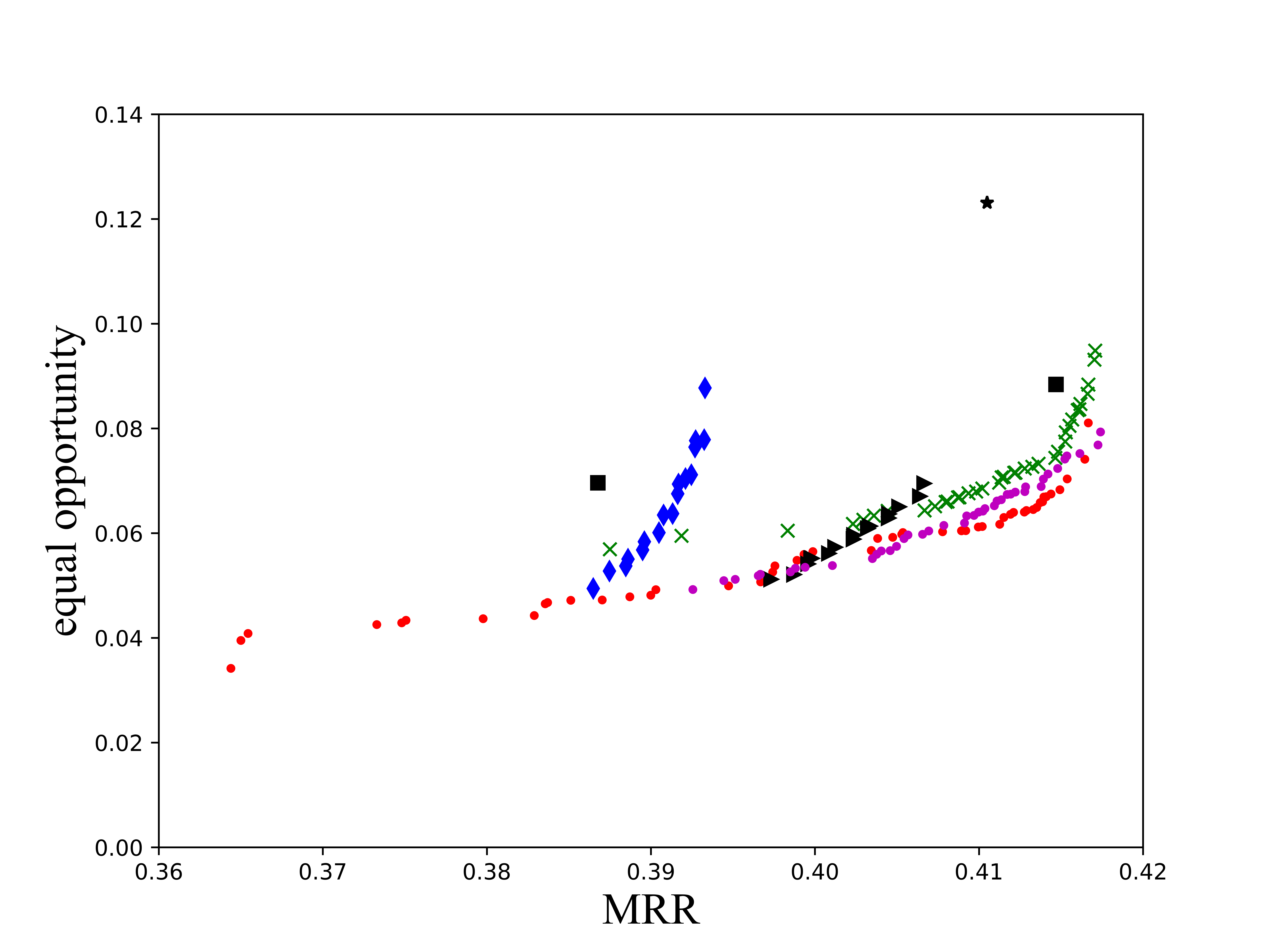}
	}
	\subfigure{
		\includegraphics[width=0.9\textwidth]{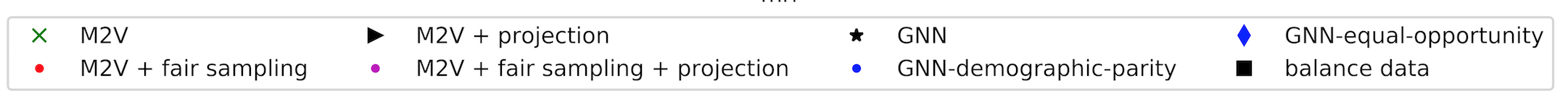}
	}
	
	\caption{The Pareto frontier of each method to demonstrate the accuracy and fairness tradeoff.  }
    \label{fig:trade-off}
\end{figure*}

\subsection{Model Selection}
Many of the methods used in our experiment have hyper-parameters that we need to tune. 
For M2V, we tune the number of walks and the length of each walk. 
For fair-sampling based methods, we tune the number of walks, the length of a walk, and the ratio $r$ in Eq. (\ref{eq:career_meta}). 
For the projection method, the hyper-parameters are the same as M2V or M2V + fair-sampling. 
For GNN-demographic-parity and GNN-equal-opportunity, the hyperparameters are $\lambda_{dp}$ and $\lambda_{eo}$ respectively. For the adversarial method, the hyperparameter is the trade-off coefficient $\gamma$ between the accuracy loss and fairness loss in the encoder. 



We used grid search to tune hyper-parameters for sampling-based methods, where the range of the number of walks and the length of each walk were $\{10,50,100,200\}$,
$r \in \{1,2,\cdots,10\}$ and the search step size was $1$.
We used grid search to tune hyper-parameters for GNNs-based methods, where $\lambda_{dp}, \lambda_{eo} \in \{10,20,\cdots,100\}$ and the search step size was $10$. 
We used grid search to tune hyper-parameters for the adversarial method, where $\gamma \in \{0.1,0.2, \cdots,1.0 \}$ and the search step size was $0.1$. 

  
  


\subsection{Experimental Results}

To study the performance of different fair representation learning algorithms for HINs, especially their ability to balance the trade-off between prediction accuracy and fairness, we plot the results on the test datasets in Figure~\ref{fig:fb_dp}-\ref{fig:ml_eo}. The Y-axis of each chart represents a fairness measure (either demographic parity in Figure~\ref{fig:fb_dp} and~\ref{fig:ml_dp} or equal opportunity in Figure~\ref{fig:fb_eo} and~\ref{fig:ml_eo}), while the X-axis represents the prediction accuracy (MRR). 
Since some algorithms such as sampling-based methods and fairness-aware GNNs-based approaches have hyper-parameters that can be tuned to achieve different fairness and accuracy tradeoffs, we show the Pareto frontier of each method which consists of models that are not dominated by other alternatives from that method. 
We say that a model $A$ dominates an alternative model $B$ if model $A$ outscores model $B$ regardless of the trade-off between fairness and accuracy -- that is, if $A$ is better than $B$ in both fairness and accuracy.  

As shown in these charts, the traditional GNN method (represented as the \emph{black star}) has relatively good MRR but poor fairness as it only tries to optimize prediction accuracy. The second baseline method that employs a naive data balancing technique to remove bias (represented as the \emph{black square}) achieved moderate prediction accuracy as well as moderate fairness. The balance-data model performed better on the MovieLens dataset than on the Facebook dataset because the Facebook dataset is more biased and thus a lot of data has to be removed from the advantaged group to achieve balance.    
Among the rest of the models, some methods such as the traditional sampling-based method (represented by the green crosses) and the fairness-aware sampling-based method (represented by red dots) are capable of achieving high MRR at the cost of low fairness (for both demographic parity and equal opportunity, the lower the value is,  the fairer it is). Other methods, such as GNN-demographic parity/equal opportunity (represented by the blue dots/blue diamonds in the charts) and a combination of fair sampling-based and projection-based methods can achieve the best fairness and a reasonably good prediction accuracy.

\begin{table*}[th!]
    \centering
        \begin{tabular}{lcccccc}
            \toprule
            Method  & $dp_{LF}$ & $dp_{MF}$ & $dp_{HF}$ & $eo_{LF}$ & $eo_{MF}$ & $eo_{HF}$ \\
            \midrule
            balance-data & 0.2973  & 0.2973 & 0.2951 & 0.2898 & 0.2661 & 0.2513 \\
            adversarial & 0.3158 & 0.3158 & 0.3158 & 0.3155 & 0.3149 & 0.2960\\
            M2V & 0.3428 & 0.3120 & 0.2951 & 0.3171 & 0.2750 & 0.2636 \\
            M2V + fair-sampling & \textbf{0.3466} & 0.3303 & 0.2856 & 0.3250 & 0.2916 &  0.2780 \\
            \midrule
            M2V + projection & 0.3177 & 0.3177 & 0.3177 & 0.3217 & 0.3088 & 0.2647 \\
            M2V + fair-sampling + projection & 0.3354 & \textbf{0.3351} & 0.3267 & 0.3284 & 0.3023 & 0.2918 \\

            \midrule
            GNN & 0.3341 & -- & -- & 0.3341 & -- & -- \\
            GNN-demographic-parity  & 0.3400 & 0.3279 & \textbf{0.3277} & -- & -- & -- \\
            GNN-equal-opportunity  & -- & -- &  -- & \textbf{0.3382} & \textbf{0.3266} & \textbf{0.3125} \\
            \bottomrule
        \end{tabular}
    \caption{Comparison of different HINs representation learning methods for career prediction on the Facebook dataset. Mean Reciprocal Rank (MRR) is reported under different demographic parity (dp)/equal opportunity (eo) constraints. We report mean of 5 runs. Here, LF, MF, and HF represent the low fairness, medium fairness and high fairness conditions. }
    \label{table:fb_result}
\end{table*}

\begin{table*}[th!]
    \centering
        \begin{tabular}{lcccccc}
           \toprule
            Method  & $dp_{LF}$ & $dp_{MF}$ & $dp_{HF}$ & $eo_{LF}$ & $eo_{MF}$ & $eo_{HF}$ \\
            \midrule
            balance-data & 0.4022 & 0.3780 & 0.3279  & 0.4088 & 0.3868 & 0.3119 \\
            adversarial & 0.3929 & 0.3922 & 0.3849 & 0.3914 & 0.3860 & 0.3734\\
            M2V & 0.4061 & 0.3840 & 0.3209 &  0.4104 & 0.3901 & 0.3160 \\
            M2V + fair-sampling  & 0.4097  & 0.3946 & 0.3265 & 0.4114 & 0.3912 & 0.3151 \\
            \midrule
            M2V + projection & 0.4064 & \textbf{0.4056} & 0.3857 & 0.4066 &  \textbf{0.3977} & 0.3356 \\
            M2V + fair-sampling + projection & \textbf{0.4143} &  0.3967 & 0.3371 & \textbf{0.4152} & {0.3949} & 0.3268 \\ 
            \midrule
            GNN & 0.4105 & -- & -- & 0.4105 & -- & -- \\
            GNN-demographic-parity  & 0.3932 & 0.3925 & \textbf{0.3881} & -- & -- & -- \\
            GNN-equal-opportunity  & -- & -- & -- & 0.3915 & 0.3886  & \textbf{0.3795} \\
            \bottomrule
        \end{tabular}
    \caption{Comparison of different HINs representation learning methods for career prediction on the MovieLens dataset. Mean Reciprocal Rank (MRR) is reported under different demographic parity (dp)/equal opportunity (eo) constraints. We report mean of 5 runs. Here, LF, MF, and HF represent the low fairness, medium fairness and high fairness conditions. }
    \label{table:ml_result}
\end{table*}

To further illustrate each model's ability in balancing the trade-off between accuracy and fairness and facilitate model comparison, we fixed the fairness dimension and only compare the prediction accuracy of these models. We choose three reference fairness thresholds to illustrate model performance under three conditions: high fairness (HF), medium fairness (MF), and low fairness (LF).  The HF condition simulates a real-world scenario where a system is used in making consequential decisions (e.g., sentencing).  Thus under this condition, we may want to choose a model with high fairness. The low fairness condition simulates a real-world scenario where the fairness of a system is not consequential (e.g., for entertainment). 

The reference fairness thresholds were based on the baseline \textit{GNN} method. 
Since traditional GNN performed poorly on the fairness dimension, we use its fairness performance as the threshold to simulate the LF condition. 
The MF condition is 75\% of demographic parity or equal opportunity value achieved by traditional GNN.
The HF condition is 50\% of demographic parity or equal opportunity value achieved by traditional GNN.
Once we select the three fairness thresholds, we report the performance of the model with the highest MRR among all the models satisfying the fairness constraint. We report the mean MRR of 5 runs. 

As shown in Table~\ref{table:fb_result}, on the Facebook dataset, under the LF condition and the demographic parity measure, traditional methods without any de-biasing such as M2V and GNN perform quite well. 
This is not surprising since the LF condition puts relatively few fairness constraints on the systems. Thus systems that totally ignore the fairness constraints (e.g., M2V and GNN) performed quite well. 
M2V + fair-sampling even outperforms traditional methods slightly. 
It demonstrates that M2V + fair-sampling does not harm accuracy under a LF condition. 
Under the MF condition and the demographic parity measure, M2V + fair-sampling and M2V + fair-sampling + projection outperform two baselines (i.e., M2V and adversarial) by 2\%. 
It demonstrates that our proposed fair-sampling methods are effective under a MF condition. 
Under the HF condition and the demographic parity measure,  GNN-demographic-parity outperforms M2V and adversarial by 3\% and 1\% respectively. 
It demonstrates that our proposed fairness-aware loss is effective under a HF condition. 
Under the equal opportunity measure, GNN-equal-opportunity is the best under all conditions. It outperforms baselines significantly. 
It demonstrates the effectiveness of fairness-aware loss under the equal opportunity measure. 
Under all fairness conditions and the equal opportunity measure, M2V + fair-sampling outperforms M2V significantly. 
It shows that the fair-sampling technique can improve M2V under all fairness conditions if the measure is equal opportunity.
In summary, under all fairness conditions, M2V + fair-sampling + projection perform well on demographic parity measure, and GNN-equal-opportunity performed well on equal opportunity measure.

The results on the MovieLens dataset are slightly different. 
First, all methods perform quite well under the LF condition. M2V + fair-sampling + projection outperforms M2V and adversarial by 1\% and 2\% respectively. 
Under the MF condition, M2V + projection outperforms M2V and adversarial by 2\% and 1\% respectively if the measure is demographic parity. 
Under the MF condition, M2V + projection outperforms M2V and adversarial by 1\% if the measure is equal opportunity. 
Under the HF condition, GNN-demographic-parity and GNN-equal-opportunity outperform M2V by 6\% on demographic parity measure and equal opportunity measure respectively. 
They are comparable with the adversarial method. 
In summary, under LF and MF fairness conditions, projection-based methods perform well. 
Under HF fairness conditions, GNN-demographic-parity and GNN-equal-opportunity methods perform well.

To summarize, based on results on Facebook and MovieLens dataset, M2V, GNN, and fair-sampling based methods can be used if under the LF condition. 
Under the HF and MF condition, project-based methods(e.g., M2V + projection and M2V + fair-sampling + projection), and GNN with fairness-aware loss methods (e.g., GNN-demographic-parity and GNN-equal-opportunity) consistently performed well.

\section{Conclusion}
In this paper, we systematically investigated a wide range of de-biasing  algorithms for HINs representation, ranging from sampling-based, projection-based, to graph neural networks (GNNs)-based approaches.  
We evaluated the effectiveness of these methods in an automated career counseling task where we mitigate harmful gender bias in career recommendation. 
Based on the evaluation results on two datasets, we identified different algorithms that are effective under different conditions, which provides valuable guidance to practitioners. 
One possible downstream application for fair HINs representation learning is communities recommendation in e-commerce platforms and social platforms. Unbiased communities recommendation relies on users' interests rather than users' sensitive attributes. We leave it to future work. 
\bibliographystyle{aaai21}
\bibliography{aaai}

\end{document}